\documentclass[
]{ceurart}

\usepackage{listings}
\usepackage{algorithm}
\usepackage{algpseudocode}
\usepackage[table]{xcolor}
\definecolor{lightcyan}{RGB}{224,255,255}
\begin{document}

\copyrightyear{2026}
\copyrightclause{Copyright for this paper by its authors.
  Use permitted under Creative Commons License Attribution 4.0
  International (CC BY 4.0).}

\conference{2nd International Workshop on Informing ML with Knowledge Engineering for Hybrid Intelligent Systems (HHAI-KEML 2026),
  July 6--7, 2026, Brussels, Belgium}

\title{Constraint-Aware Counterfactual Editing for Aspect-Based Sentiment Analysis}

\author[1]{S M Rafiuddin}
\ead{srafiud@okstate.edu}

\author[1]{Vamsi Krishna Pavuluri}

\author[1]{Atriya Sen}

\address[1]{Department of Computer Science,
Oklahoma State University,
Stillwater, Oklahoma, USA}


\begin{abstract}
Aspect-Based Sentiment Analysis (ABSA) requires models to identify sentiment toward specific aspects rather than relying on the global polarity of a sentence. This makes counterfactual evaluation especially challenging: a valid counterfactual should flip the sentiment of one target aspect while preserving the sentiment of all non-target aspects, semantic meaning, fluency, and factual consistency. Existing counterfactual generation methods often focus on sentence-level label flipping and may produce edits that are fluent but aspect-invalid, semantically drifting, or contradictory. To address this limitation, we propose \textbf{CAVE-ABSA}, a \textbf{C}onstraint-\textbf{A}ware \textbf{V}alidated \textbf{E}diting framework for generating and validating aspect-level counterfactuals. CAVE-ABSA localizes the opinion span associated with the target aspect, performs controlled counterfactual rewriting, refines candidates through a repair module, and filters them using aspect-level verification, semantic similarity, AMR-guided structural preservation, edit minimality, fluency, and contradiction detection. The framework is designed to construct validated counterfactual ABSA datasets for robustness evaluation and data augmentation. By explicitly separating generation from validation, CAVE-ABSA provides a principled approach for producing meaningful aspect-local counterfactuals and for testing whether ABSA models truly rely on aspect-grounded sentiment reasoning.
\end{abstract}


\maketitle

\section{Introduction}

\textbf{\textit{Aspect-Based Sentiment Analysis (ABSA)}} identifies sentiment polarity toward specific aspects or opinion targets rather than assigning a single global label to a sentence~\cite{pontiki2014semeval,pontiki2015semeval,pontiki2016semeval}. This distinction is important because a sentence may express different sentiments toward different aspects, e.g., ``The battery is excellent, but the screen is dull.'' Although \textbf{\textit{pretrained language models}} have improved ABSA performance by modeling aspect-context interactions~\cite{devlin2019bert,sun2019utilizing,li2021solving,tian2024context}, strong benchmark results do not necessarily imply robust \textbf{\textit{aspect-grounded reasoning}}. Models may still exploit global sentiment, polarity shortcuts, or dataset artifacts instead of correctly grounding sentiment in the target aspect~\cite{jiang2019challenge}.

\textbf{\textit{Counterfactual examples}} offer a useful way to test such robustness. Prior work on counterfactually augmented data shows that minimally edited examples can reduce reliance on spurious correlations and reveal brittle decision boundaries~\cite{kaushik2020learning,gardner2020evaluating,sen2021counterfactually,joshi2022ineffectiveness}. However, most counterfactual generation methods are designed for sentence-level classification, where changing the global label is sufficient~\cite{ross2021explaining,wu2021polyjuice}. This assumption does not hold for ABSA: a valid counterfactual must flip the sentiment of one target aspect while preserving all non-target aspect sentiments.

We frame this requirement as an \textbf{\textit{aspect-local counterfactual editing}} problem. Given an ABSA instance with a sentence, target aspect, and original sentiment label, the goal is to generate a minimally edited, fluent, and meaningful counterfactual where only the target aspect sentiment changes. Simple polarity-changing perturbations are often insufficient. For example, changing ``The trackpad is nice and smooth'' to ``The trackpad is poor and smooth'' introduces mixed polarity, while changing ``easy to use'' to ``a bad laptop'' changes global sentiment without directly flipping the target aspect. Such cases show that ABSA counterfactuals require \textbf{\textit{aspect-level verification}} rather than sentence-level sentiment change alone.

Recent ABSA-specific work identifies aspect-related opinion expressions and generates reversed-polarity replacements~\cite{wu2024novel}. While promising, meaningful counterfactual generation remains challenging: lexical substitutions can be unnatural, masked language models may produce label-invalid edits, and direct semantic-graph edits can introduce grammatical errors or semantic drift. CEval further shows that counterfactual generation often trades off label-flipping success against fluency, coherence, and minimality~\cite{nguyen2024ceval}.

In this paper, we propose \textbf{\textit{CAVE-ABSA}}, a \textbf{C}onstraint-\textbf{A}ware \textbf{V}alidated \textbf{E}diting framework for ABSA counterfactual generation. CAVE-ABSA localizes the opinion span associated with the target aspect, performs controlled phrase- or clause-level rewriting, refines candidates through a repair step, and validates them using \textbf{\textit{aspect-level flipping}}, \textbf{\textit{non-target preservation}}, \textbf{\textit{semantic similarity}}, \textbf{\textit{fluency}}, \textbf{\textit{minimality}}, \textbf{\textit{contradiction filtering}}, and \textbf{\textit{AMR-guided structural preservation}}~\cite{sundararajan2017axiomatic,raffel2020exploring,banarescu2013abstract,reimers2019sentence}. The framework can be used to construct validated counterfactual ABSA datasets for robustness evaluation and to augment training data with high-quality minimally paired examples.

Our contributions are: (1) We formalize \textbf{\textit{ABSA counterfactual editing}} as a \textbf{\textit{constrained generation problem}} requiring target aspect flipping, non-target preservation, semantic consistency, fluency, minimality, and contradiction avoidance; (2) We propose \textbf{\textit{CAVE-ABSA}}, a \textbf{\textit{hybrid generation-and-verification framework}} based on opinion-span localization, controlled rewriting, repair, AMR-guided structure checking, and ABSA-specific validation; (3) We introduce \textbf{\textit{evaluation criteria}} for measuring counterfactual validity, non-target consistency, semantic preservation, edit minimality, fluency, contradiction rate, and (4) We position \textbf{\textit{validated counterfactuals}} as a resource for both ABSA robustness evaluation and data augmentation.

\section{Related Work}
\label{sec:related_work}

\textbf{\textit{Aspect-Based Sentiment Analysis (ABSA)}} identifies sentiment polarity toward specific aspects rather than assigning a single global label to a sentence~\cite{pontiki2014semeval,pontiki2015semeval,pontiki2016semeval}. SemEval ABSA benchmarks established the standard tasks of aspect extraction and aspect-level polarity classification, while MAMS highlighted the challenge of multi-aspect sentences with different polarities in the same context~\cite{jiang2019challenge}. Pretrained language models have improved ABSA by modeling aspect-context interactions through auxiliary-sentence classification and text-to-text formulations~\cite{devlin2019bert,sun2019utilizing,li2021solving,tian2024context}; however, high benchmark performance does not guarantee \textbf{\textit{aspect-grounded reasoning}}, as models may still rely on global sentiment, polarity shortcuts, or dataset artifacts~\cite{jiang2019challenge}. \textbf{\textit{Counterfactual data augmentation}} has been used to reduce spurious correlations and improve robustness by minimally editing examples so that their labels change~\cite{kaushik2020learning,gardner2020evaluating}, although subsequent studies show that the resulting gains are not always consistent~\cite{sen2021counterfactually,joshi2022ineffectiveness}. Automatic methods such as MiCE and Polyjuice generate minimal or controllable counterfactual edits for explanation, evaluation, and augmentation~\cite{ross2021explaining,wu2021polyjuice}, while recent LLM-based approaches such as CORE and DISCO improve scalability through retrieve-then-edit generation and over-generation followed by filtering~\cite{dixit2022core,chen2023disco}. Nevertheless, these methods mainly target sentence-level label flipping and do not explicitly enforce \textbf{\textit{aspect-specific preservation}}. ABSA counterfactual editing is more constrained: a valid edit must flip the sentiment of the target aspect while preserving all non-target aspect sentiments. Wu et al.~\cite{wu2024novel} address this setting by using integrated gradients to identify aspect-related opinion expressions and T5 to generate reversed-polarity expressions; however, existing ABSA counterfactual methods mainly emphasize augmentation and downstream performance, while broader validity constraints such as semantic consistency, fluency, minimality, contradiction avoidance, and non-target preservation are not jointly enforced. Our preliminary exploration also shows that lexical replacement, masked language model substitution, and direct semantic-graph editing often produce unnatural, contradictory, or target-mismatched counterfactuals. Evaluating counterfactuals therefore requires more than checking whether a label changes. CEval emphasizes that counterfactual generation should be evaluated using both validity and text-quality criteria, including fluency, coherence, and semantic preservation~\cite{nguyen2024ceval}. Embedding-based similarity methods such as Sentence-BERT can estimate semantic closeness~\cite{reimers2019sentence}, but may miss structural drift. Therefore, we use \textbf{\textit{Abstract Meaning Representation (AMR)}} as an auxiliary semantic-structure signal, since AMR represents sentence meaning as a rooted labeled graph~\cite{banarescu2013abstract}. Unlike direct AMR generation, we use AMR as a verification layer to support semantic preservation. Overall, \textbf{\textit{CAVE-ABSA}} differs from prior work by framing counterfactual ABSA as a \textbf{\textit{constraint-aware generation and validation}} problem, enforcing target aspect flipping, non-target preservation, semantic consistency, fluency, minimality, contradiction avoidance, and AMR-guided structure checking.

\section{Problem Formulation}
\label{sec:problem_formulation}

We formulate \textbf{\textit{counterfactual editing}} for \textbf{\textit{Aspect-Based Sentiment Analysis (ABSA)}} as a constrained generation problem. Given an input sentence $x$, let $A(x)=\{a_1,a_2,\ldots,a_m\}$ denote the set of annotated aspect terms in the sentence. Each aspect $a_j \in A(x)$ is associated with a sentiment polarity label $y_j \in \mathcal{Y}$, where $\mathcal{Y}$ denotes the task-specific sentiment label space, such as positive, negative, and neutral. For a target aspect $a^\ast \in A(x)$ with original sentiment label $y^\ast$, the goal is to generate a counterfactual sentence $x'$ that changes the sentiment toward $a^\ast$ while preserving the sentiment toward all non-target aspects. Let $f_{\theta}(x,a)$ denote an ABSA classifier that predicts the sentiment polarity of aspect $a$ in sentence $x$. A valid counterfactual sentence $x'$ should first satisfy the \textbf{\textit{target-aspect flip constraint}} $f_{\theta}(x',a^\ast)=y^{cf}$, where $y^{cf}\neq y^\ast$. In the binary setting, $y^{cf}$ is defined as the opposite polarity of $y^\ast$, i.e., $y^{cf}=\text{negative}$ if $y^\ast=\text{positive}$, and $y^{cf}=\text{positive}$ if $y^\ast=\text{negative}$. However, target flipping alone is not sufficient for ABSA. If the original sentence contains multiple aspects, a counterfactual edit should not alter the sentiment of non-target aspects. Therefore, for every $a_j \in A(x)$ where $a_j \neq a^\ast$, the \textbf{\textit{preservation constraint}} must hold: $f_{\theta}(x',a_j)=y_j,\ \forall a_j \in A(x),\ a_j \neq a^\ast$. In addition to aspect-level label constraints, the edited sentence should remain close to the original sentence in meaning and form. We therefore require \textbf{\textit{semantic preservation}}, \textbf{\textit{edit minimality}}, \textbf{\textit{fluency}}, and \textbf{\textit{contradiction avoidance}}, expressed as $\mathrm{Sim}(x,x') \geq \tau_{\mathrm{sem}}$, $\mathrm{EditDistance}(x,x') \leq \delta$, $\mathrm{Fluency}(x') \geq \tau_{\mathrm{flu}}$, and $\mathrm{Contradiction}(x')=0$. Here, $\mathrm{Sim}(x,x')$ measures semantic similarity between the original and edited sentences, $\mathrm{EditDistance}(x,x')$ measures the magnitude of the edit, $\mathrm{Fluency}(x')$ estimates grammaticality and naturalness, and $\mathrm{Contradiction}(x')$ indicates whether the candidate introduces conflicting sentiment cues or inconsistent facts. These constraints are necessary because a sentence can flip a classifier prediction while still being invalid as a meaningful counterfactual. For example, ``the trackpad is poor and smooth'' contains a negative cue, but it also preserves a positive cue for the same aspect, producing a mixed and contradictory sentiment expression. We define the \textbf{\textit{valid counterfactual set}} for an ABSA instance $(x,a^\ast,y^\ast)$ as $\mathcal{V}(x,a^\ast)=\{x' : f_{\theta}(x',a^\ast)=y^{cf},\ y^{cf}\neq y^\ast;\ f_{\theta}(x',a_j)=y_j,\ \forall a_j\neq a^\ast;\ \mathrm{Sim}(x,x') \geq \tau_{\mathrm{sem}};\ \mathrm{EditDistance}(x,x') \leq \delta;\ \mathrm{Fluency}(x') \geq \tau_{\mathrm{flu}};\ \mathrm{Contradiction}(x')=0\}$. The objective of \textbf{\textit{CAVE-ABSA}} is to generate a small set of high-quality counterfactual candidates $\mathcal{C}(x,a^\ast)=\{x'_1,x'_2,\ldots,x'_K\}$, where $x'_k \in \mathcal{V}(x,a^\ast)$ and each $x'_k$ provides a fluent, meaningful, and minimally edited alternative to the original sentence. To rank candidate counterfactuals, we use a \textbf{\textit{composite score}} that combines target flipping, non-target preservation, semantic similarity, fluency, minimality, structural preservation, and contradiction penalty: $S(x')=\lambda_1 P_{\mathrm{flip}}(x')+\lambda_2 P_{\mathrm{keep}}(x')+\lambda_3 S_{\mathrm{sem}}(x,x')+\lambda_4 S_{\mathrm{flu}}(x')+\lambda_5 S_{\mathrm{edit}}(x,x')+\lambda_6 S_{\mathrm{struct}}(x,x')-\lambda_7 C(x')$. Here, $P_{\mathrm{flip}}(x')$ measures whether the target aspect sentiment is flipped, $P_{\mathrm{keep}}(x')$ measures whether non-target aspect sentiments are preserved, $S_{\mathrm{sem}}(x,x')$ measures semantic similarity, $S_{\mathrm{flu}}(x')$ measures fluency, $S_{\mathrm{edit}}(x,x')$ rewards minimal edits, $S_{\mathrm{struct}}(x,x')$ measures semantic-structure preservation, and $C(x')$ penalizes contradiction. The final counterfactual set is obtained by selecting the top-$K$ candidates that satisfy all hard validity constraints.

\section{Methodology}
\label{sec:method}

\begin{figure}[t]
    \centering
    \includegraphics[width=\linewidth]{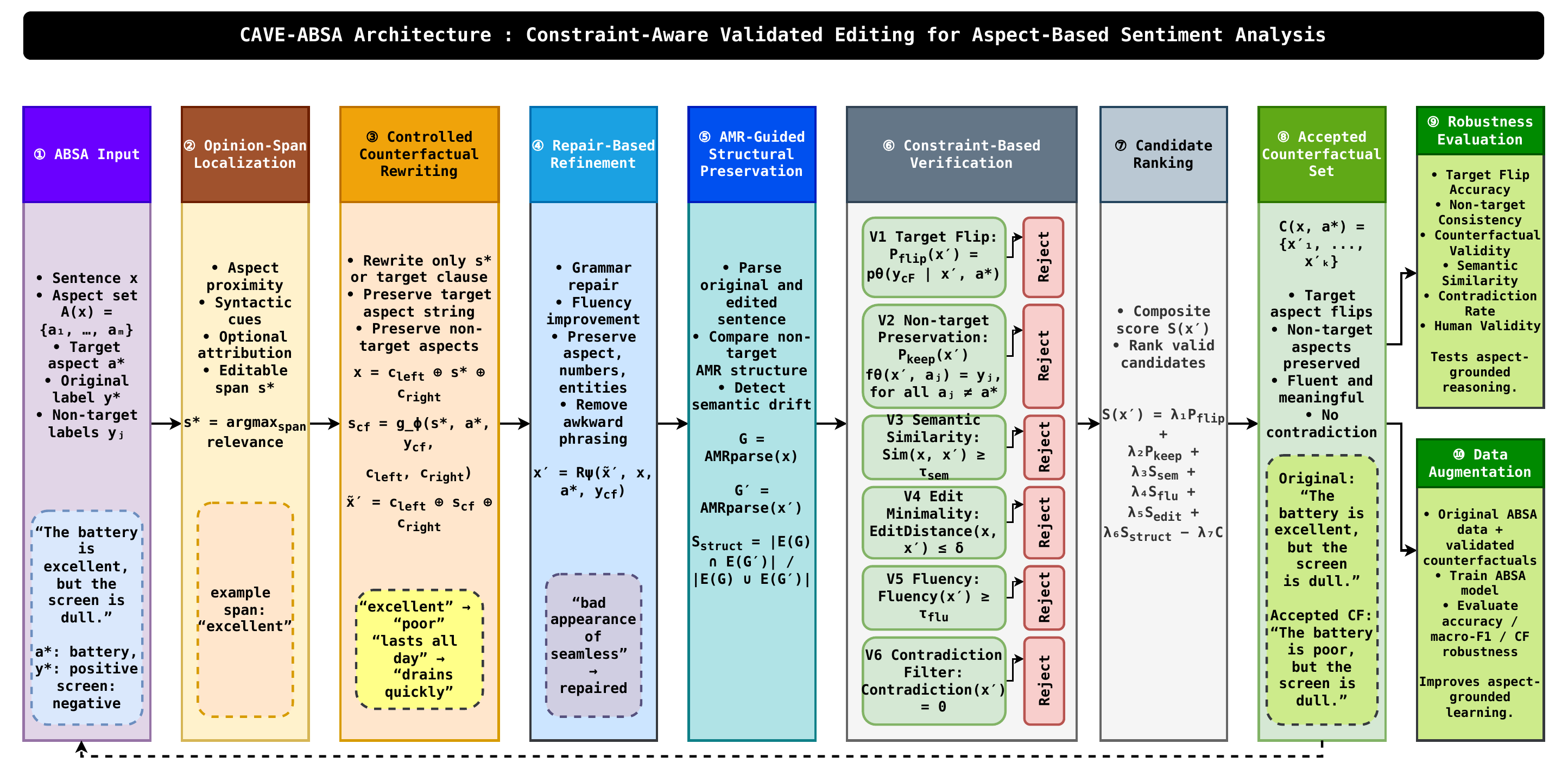}
\caption{
\textbf{\textit{Overview of the CAVE-ABSA architecture.}}
Given an ABSA instance consisting of an input sentence $x$, an aspect set $A(x)$, a target aspect $a^\ast$, its original sentiment label $y^\ast$, and non-target aspect labels $y_j$, \textbf{\textit{CAVE-ABSA}} generates validated aspect-level counterfactuals through a left-to-right pipeline. 
The framework first performs \textbf{\textit{opinion-span localization}} to identify the editable span $s^\ast$ associated with the target aspect. 
It then applies \textbf{\textit{controlled counterfactual rewriting}} to modify only the target opinion span or clause while preserving the target aspect string, non-target aspects, and factual context. 
The generated candidate is refined through a \textbf{\textit{repair module}} to improve grammar and fluency, followed by \textbf{\textit{AMR-guided structural preservation}} to detect semantic drift in non-target content. 
A \textbf{\textit{constraint-based verification}} module filters candidates using six hard validity checks: target aspect flipping, non-target preservation, semantic similarity, edit minimality, fluency, and contradiction avoidance. 
Valid candidates are ranked using a \textbf{\textit{composite score}} that combines target flip confidence, non-target consistency, semantic similarity, fluency, edit minimality, AMR structural preservation, and contradiction penalty. 
The final accepted counterfactual set $\mathcal{C}(x,a^\ast)$ is used for \textbf{\textit{robustness evaluation}} and \textbf{\textit{data augmentation}}, enabling tests of whether ABSA models rely on aspect-grounded sentiment reasoning rather than global sentence polarity.
}
    \label{fig:cave_absa_architecture}
\end{figure}

We propose \textbf{CAVE-ABSA}, a \textbf{C}onstraint-\textbf{A}ware \textbf{V}alidated \textbf{E}diting framework for generating aspect-local counterfactuals for ABSA. Given an input sentence $x$, a target aspect $a^\ast$, and its original polarity $y^\ast$, CAVE-ABSA generates a set of candidate counterfactuals and retains only those that satisfy target-aspect flipping, non-target aspect preservation, semantic consistency, fluency, minimality, and contradiction-avoidance constraints. As shown in Figure~\ref{fig:cave_absa_architecture}, the framework consists of five main stages: opinion-span localization, controlled counterfactual rewriting, repair-based refinement, semantic-structure checking, and constraint-based verification. The verified candidates are then ranked and used to construct the accepted counterfactual set for robustness evaluation and data augmentation.

\subsection{Overview}

The overall \textbf{\textit{CAVE-ABSA pipeline}} is represented as $x \rightarrow \text{\textbf{\textit{Opinion Span Localization}}} \rightarrow \text{\textbf{\textit{Controlled Rewriting}}} \rightarrow \text{\textbf{\textit{Repair}}} \rightarrow \text{\textbf{\textit{Verification}}} \rightarrow \mathcal{C}(x,a^\ast)$. Unlike direct \textbf{\textit{lexical substitution}} or direct \textbf{\textit{AMR concept replacement}}, CAVE-ABSA does not treat counterfactual editing as a single-word replacement problem. Instead, it identifies the \textbf{\textit{opinion-bearing span}} associated with the target aspect and rewrites that span or its surrounding clause under explicit \textbf{\textit{aspect-level constraints}}. This design is motivated by the observation that aspect sentiment is often expressed through short opinion phrases such as ``lasts all day,'' ``easy to use,'' or ``poor build quality,'' rather than through isolated polarity words.

\subsection{Opinion-Span Localization}

For each ABSA instance $(x,a^\ast,y^\ast)$, we first identify the local \textbf{\textit{opinion span}} $s^\ast$ that expresses sentiment toward the target aspect $a^\ast$. This span acts as the \textbf{\textit{editable region}}. We use a hybrid localization strategy combining \textbf{\textit{aspect proximity}}, \textbf{\textit{syntactic cues}}, and optional \textbf{\textit{model attribution}}. Let $T(x)=\{t_1,t_2,\ldots,t_n\}$ denote the token sequence of $x$. We assign each token $t_i$ a relevance score with respect to the target aspect as $r_i = \alpha_1 r_i^{\mathrm{prox}} + \alpha_2 r_i^{\mathrm{dep}} + \alpha_3 r_i^{\mathrm{attr}}$, where $r_i^{\mathrm{prox}}$ measures distance from the aspect term, $r_i^{\mathrm{dep}}$ measures dependency connectivity to the aspect, and $r_i^{\mathrm{attr}}$ is an optional attribution score from an ABSA classifier, such as integrated gradients~\cite{sundararajan2017axiomatic,wu2024novel}. The highest-scoring contiguous region is selected as the opinion span, $s^\ast = \arg\max_{s \subseteq x} \sum_{t_i \in s} r_i$. When attribution scores are unavailable, \textbf{\textit{CAVE-ABSA}} falls back to dependency- and clause-based localization. Specifically, it selects the clause containing the target aspect and identifies adjectives, adverbs, sentiment-bearing verbs, and adjectival complements near the aspect as candidate opinion expressions.

\subsection{Controlled Counterfactual Rewriting}

After identifying the target \textbf{\textit{opinion span}} $s^\ast$, CAVE-ABSA generates counterfactual candidates by rewriting only the target span or target clause. Let $x$ be decomposed into three parts as $x = c_{\mathrm{left}} \oplus s^\ast \oplus c_{\mathrm{right}}$, where $c_{\mathrm{left}}$ and $c_{\mathrm{right}}$ are the preserved context segments. The rewriting module generates a counterfactual opinion span $s^{cf}$ with the desired polarity $y^{cf}$ as $s^{cf} = g_{\phi}(s^\ast, a^\ast, y^{cf}, c_{\mathrm{left}}, c_{\mathrm{right}})$. The candidate sentence is then reconstructed as $\tilde{x}' = c_{\mathrm{left}} \oplus s^{cf} \oplus c_{\mathrm{right}}$. The generator is prompted or constrained to follow four rules: preserve the target aspect string, change only the sentiment toward the target aspect, keep all non-target aspects and factual details unchanged, and return a fluent complete sentence. This \textbf{\textit{span-level rewriting}} is more flexible than \textbf{\textit{antonym replacement}} because it can transform expressions such as ``lasts all day'' into ``drains quickly,'' or ``easy to use'' into ``difficult to use,'' rather than replacing a single token.
\subsection{Repair-Based Refinement}

Generated candidates may satisfy the desired polarity change but still contain grammatical errors or unnatural phrasing. Therefore, each raw candidate $\tilde{x}'$ is passed through a \textbf{\textit{repair module}} as $x' = R_{\psi}(\tilde{x}', x, a^\ast, y^{cf})$, where $R_{\psi}$ rewrites the candidate for grammar and fluency while preserving the intended counterfactual meaning. The repair step is explicitly constrained not to change the target aspect, non-target aspects, numbers, named entities, or product-specific facts. This step is important because many automatic edits produce malformed sentences such as ``bad appearance of seamless'' or contradictory expressions such as ``poor and smooth.'' The repair module attempts to convert such outputs into grammatical alternatives, but candidates are still rejected if they fail the final \textbf{\textit{verification stage}}.

\subsection{AMR-Guided Structural Preservation}

To reduce \textbf{\textit{semantic drift}}, CAVE-ABSA uses \textbf{\textit{Abstract Meaning Representation (AMR)}} as a structure-level preservation signal~\cite{banarescu2013abstract}. We parse both the original sentence and the candidate counterfactual into AMR graphs as $G = \mathrm{AMRParse}(x)$ and $G' = \mathrm{AMRParse}(x')$. Since the target opinion subgraph is expected to change, we exclude the localized target region from the structural comparison. Let $E(G)$ and $E(G')$ denote the sets of non-target semantic edges in the original and edited AMR graphs. We compute \textbf{\textit{structural preservation}} as $S_{\mathrm{struct}}(x,x') = \frac{|E(G)\cap E(G')|}{|E(G)\cup E(G')|}$. This score helps reject candidates that remain lexically similar but alter important non-target relations. For example, if an edit flips the target aspect but also changes a comparison, removes a causal relation, or changes the relation between a product and its attribute, its structural preservation score decreases.

\subsection{Constraint-Based Verification}

The \textbf{\textit{verification module}} is the core component of CAVE-ABSA. It ensures that a generated sentence is not accepted merely because it contains an opposite-polarity word. Each candidate $x'$ is evaluated using six constraints. First, the target aspect must flip, measured as $P_{\mathrm{flip}}(x') = p_{\theta}(y^{cf}\mid x',a^\ast)$. Second, all non-target aspects must preserve their original labels, measured as $P_{\mathrm{keep}}(x') = \frac{1}{|A(x)\setminus\{a^\ast\}|}\sum_{a_j \neq a^\ast} p_{\theta}(y_j\mid x',a_j)$; for single-aspect sentences, we set $P_{\mathrm{keep}}(x')=1$. Third, the candidate must remain semantically close to the original sentence, computed as $S_{\mathrm{sem}}(x,x') = \cos(h(x),h(x'))$, where $h(\cdot)$ is a sentence embedding model such as Sentence-BERT~\cite{reimers2019sentence}. Fourth, the edit should be minimal, measured as $S_{\mathrm{edit}}(x,x') = 1 - \frac{d(x,x')}{|x|}$, where $d(x,x')$ is a token-level edit distance. Fifth, the sentence must be fluent, measured as $S_{\mathrm{flu}}(x') = \exp(-\lambda \cdot \mathrm{PPL}(x'))$, where lower perplexity indicates better fluency. Finally, the candidate must not contain contradictory or mixed-polarity expressions, represented as $C(x')=1$ if contradiction is detected and $C(x')=0$ otherwise. \textbf{\textit{Contradiction detection}} combines local polarity checks, rule-based mixed-cue detection, and optionally a natural language inference model. For example, expressions such as ``poor and smooth,'' ``terrible but excellent,'' or ``sluggish and easy to use'' are treated as invalid when they refer to the same target aspect.

\subsection{Candidate Ranking and Selection}

All valid candidates are ranked using the \textbf{\textit{composite score}} introduced in Section~\ref{sec:problem_formulation}. In practice, we first apply hard constraints: $P_{\mathrm{flip}}(x') \geq \tau_{\mathrm{flip}}$, $P_{\mathrm{keep}}(x') \geq \tau_{\mathrm{keep}}$, $S_{\mathrm{sem}}(x,x') \geq \tau_{\mathrm{sem}}$, $S_{\mathrm{flu}}(x') \geq \tau_{\mathrm{flu}}$, and $C(x')=0$. Candidates that fail any hard constraint are discarded. The remaining candidates are ranked by $S(x') = \lambda_1 P_{\mathrm{flip}}(x') + \lambda_2 P_{\mathrm{keep}}(x') + \lambda_3 S_{\mathrm{sem}}(x,x') + \lambda_4 S_{\mathrm{flu}}(x') + \lambda_5 S_{\mathrm{edit}}(x,x') + \lambda_6 S_{\mathrm{struct}}(x,x') - \lambda_7 C(x')$. The top-$K$ candidates are selected as the final \textbf{\textit{counterfactual set}}, $\mathcal{C}(x,a^\ast)=\operatorname{TopK}_{x' \in \mathcal{V}(x,a^\ast)} S(x')$.

\subsection{Algorithm} 
Algorithm~\ref{alg:cave_absa} summarizes the \textbf{\textit{CAVE-ABSA}} pipeline. 

\begin{algorithm}[t] 
\caption{\textbf{\textit{CAVE-ABSA Counterfactual Editing}}} 
\label{alg:cave_absa} 
\begin{algorithmic}[1] 
\Require Sentence $x$, target aspect $a^\ast$, original polarity $y^\ast$, aspect set $A(x)$ 
\Ensure Valid counterfactual set $\mathcal{C}(x,a^\ast)$ 
\State Determine target polarity $y^{cf}$ such that $y^{cf}\neq y^\ast$ 
\State Localize target \textbf{\textit{opinion span}} $s^\ast$ associated with $a^\ast$ 
\State Parse original sentence into \textbf{\textit{AMR graph}} $G$ 
\State Initialize candidate set $\mathcal{U}\leftarrow \emptyset$ 
\For{$i=1$ to $N$} 
    \State Generate raw candidate $\tilde{x}'_i$ by \textbf{\textit{controlled rewriting}} of $s^\ast$ 
    \State Repair $\tilde{x}'_i$ to obtain fluent candidate $x'_i$ 
    \State Compute target flip score $P_{\mathrm{flip}}(x'_i)$ 
    \State Compute non-target preservation score $P_{\mathrm{keep}}(x'_i)$ 
    \State Compute semantic, fluency, edit, and AMR-structure scores 
    \State Detect contradiction $C(x'_i)$ 
    \If{$x'_i$ satisfies all \textbf{\textit{validity constraints}}} 
        \State Add $x'_i$ to $\mathcal{U}$ 
    \EndIf 
\EndFor 
\State Rank candidates in $\mathcal{U}$ using $S(x')$ 
\State \Return top-$K$ candidates as $\mathcal{C}(x,a^\ast)$ 
\end{algorithmic} 
\end{algorithm}


\section{Experimental Setup}
\label{sec:experiments}

We evaluate \textbf{\textit{CAVE-ABSA}} from two complementary perspectives: (i) the \textbf{\textit{quality of generated counterfactual examples}}, and (ii) their usefulness for \textbf{\textit{evaluating and improving ABSA models}}. The first setting measures whether the generated sentence is a valid and meaningful \textbf{\textit{aspect-level counterfactual}}. The second setting tests whether validated counterfactuals can expose model brittleness and improve \textbf{\textit{aspect-grounded sentiment classification}}.

\subsection{Datasets}

We conduct experiments on standard \textbf{\textit{ABSA benchmarks}}. Our primary dataset is the \textbf{\textit{SemEval-2014 Laptop}} domain, which contains product reviews annotated with aspect terms and sentiment polarity labels~\cite{pontiki2014semeval}. We also consider the \textbf{\textit{SemEval-2014 Restaurant}} domain and \textbf{\textit{MAMS}} for additional evaluation. MAMS is especially useful because it contains multi-aspect sentences where different aspects can have different sentiment polarities~\cite{jiang2019challenge}. This setting directly tests whether a counterfactual generator can flip one target aspect while preserving non-target aspects. For each dataset, we construct counterfactual candidates only for instances with positive or negative target polarities. Neutral instances are excluded from polarity-flip generation because the desired counterfactual direction is less clearly defined. For each eligible instance $(x,a^\ast,y^\ast)$, we generate up to $K$ counterfactual candidates.

\subsection{Baselines}

We compare \textbf{\textit{CAVE-ABSA}} against representative counterfactual generation baselines. \textbf{\textit{Lexicon Replacement}} replaces sentiment-bearing words near the target aspect with opposite-polarity words using an antonym or sentiment lexicon; it is simple and interpretable, but often produces unnatural or contradictory outputs. \textbf{\textit{Masked Language Model Replacement}} masks opinion words near the target aspect and uses a masked language model to generate replacements; it can produce locally fluent substitutions, but does not explicitly guarantee aspect-level polarity flipping. \textbf{\textit{Prompt-only Rewriting}} uses a sequence-to-sequence generator or instruction-tuned model to rewrite the sentence so that the target aspect polarity changes, but does not include the full verification pipeline. \textbf{\textit{Direct AMR Concept Editing}} parses the original sentence into an AMR graph, replaces sentiment-bearing concepts with opposite-polarity concepts, and generates text back from the edited graph, testing whether semantic-graph editing alone is sufficient. \textbf{\textit{Attribution-guided T5 Rewriting}} follows the idea of identifying important opinion expressions and generating reversed-polarity expressions using a text-to-text model~\cite{wu2024novel}. \textbf{\textit{CAVE-ABSA}} is our full method, combining opinion-span localization, controlled rewriting, repair, semantic-structure checking, and constraint-based verification.

\subsection{Automatic Evaluation Metrics}

We evaluate each generated counterfactual using both \textbf{\textit{aspect-level validity metrics}} and \textbf{\textit{text-quality metrics}}. \textbf{\textit{Target Flip Accuracy}} measures the percentage of candidates for which the predicted sentiment of the target aspect changes to the desired counterfactual polarity. \textbf{\textit{Non-target Consistency}} measures the percentage of candidates for which all non-target aspect polarities remain unchanged. \textbf{\textit{Counterfactual Validity}} measures the percentage of candidates satisfying both target flip and non-target preservation constraints. \textbf{\textit{Semantic Similarity}} measures the semantic closeness between the original and edited sentence, computed using sentence embeddings such as Sentence-BERT~\cite{reimers2019sentence}. \textbf{\textit{Edit Minimality}} is computed using normalized token-level edit distance between $x$ and $x'$, where lower edit distance indicates a more minimal counterfactual. \textbf{\textit{Fluency}} measures grammaticality and naturalness using a language-model-based fluency score or perplexity. \textbf{\textit{Post-hoc Contradiction Rate}} measures the percentage of retained candidates that are later judged to contain mixed-polarity or contradictory expressions, such as ``poor and smooth'' for the same aspect, by an external evaluator after automatic filtering. This metric is distinct from the internal contradiction filter used during CAVE-ABSA verification; therefore, a non-zero value indicates residual contradictions missed by the automatic verifier. \textbf{\textit{AMR Structural Preservation}} measures the overlap between the non-target semantic structures of the original and edited sentences, indicating whether the candidate preserves sentence-level meaning beyond surface similarity. Unless otherwise stated, quality metrics are computed over retained candidates that pass the filtering stage, whereas \textbf{\textit{Generation Yield}} is computed over all initially generated candidates before filtering.

\subsection{Downstream ABSA Evaluation}

To evaluate whether generated counterfactuals are useful for training, we augment the original ABSA training set with \textbf{\textit{validated counterfactual examples}}. We compare models trained under five settings: (1) original training data only; (2) original data plus \textbf{\textit{lexicon-based counterfactuals}}; (3) original data plus \textbf{\textit{prompt-only counterfactuals}}; (4) original data plus \textbf{\textit{direct AMR-edited counterfactuals}}; and (5) original data plus \textbf{\textit{CAVE-ABSA counterfactuals}}. We evaluate downstream ABSA performance using \textbf{\textit{accuracy}} and \textbf{\textit{macro-F1}}. We also evaluate robustness on counterfactual test pairs, where a model is expected to change its prediction for the edited target aspect while preserving predictions for non-target aspects.

\subsection{Ablation Study}

We perform ablation studies to measure the contribution of each component: (1) \textbf{\textit{Without opinion-span localization}}, where the full sentence is rewritten instead of the target opinion span; (2) \textbf{\textit{Without repair}}, where the fluency-improvement step is removed; (3) \textbf{\textit{Without contradiction filtering}}, where mixed-polarity outputs are allowed; (4) \textbf{\textit{Without AMR structural checking}}, where only embedding similarity is used for semantic preservation; (5) \textbf{\textit{Without non-target verification}}, where only target aspect flipping is enforced; and (6) \textbf{\textit{Generic sentiment verifier}}, where the ABSA verifier is replaced with a sentence-level sentiment classifier. These ablations test whether each constraint is necessary for producing valid and meaningful ABSA counterfactuals.


\section{Results and Analysis}
\label{sec:results}

This section reports the effectiveness of CAVE-ABSA in generating valid and meaningful aspect-level counterfactuals. We analyze the results from three perspectives: counterfactual generation quality, usefulness for ABSA robustness evaluation, and the contribution of each framework component.

\subsection{Counterfactual Generation Quality}

Table~\ref{tab:generation_quality} compares \textbf{CAVE-ABSA} with representative counterfactual generation baselines. We evaluate generation quality using six criteria: \textbf{\textit{Target Flip}}, \textbf{\textit{Non-target Keep}}, \textbf{\textit{Semantic Similarity}}, \textbf{\textit{Fluency}}, \textbf{\textit{Post-hoc Contradiction Rate}}, and \textbf{\textit{Generation Yield}}. Target Flip measures whether the edited aspect changes to the desired counterfactual polarity, while Non-target Keep measures whether the sentiment labels of unchanged aspects remain stable. Semantic Similarity and Fluency capture meaning preservation and linguistic naturalness. Post-hoc Contradiction Rate measures residual mixed-polarity or inconsistent generations detected after automatic filtering. All quality metrics are computed over retained candidates, while Generation Yield is computed over all initially generated candidates before filtering.

The results show that \textbf{CAVE-ABSA} achieves the best overall balance across validity, quality, and filtering efficiency. It obtains the highest \textbf{\textit{Target Flip}} score of \textbf{92.4}, the highest \textbf{\textit{Non-target Keep}} score of \textbf{91.1}, and the strongest \textbf{\textit{Semantic Similarity}} score of \textbf{89.3}. It also achieves the highest Fluency score of \textbf{91.8} while reducing the Contradiction Rate to only \textbf{4.6}, substantially lower than all baselines. These results indicate that CAVE-ABSA does not merely generate fluent polarity-changing sentences; it produces counterfactuals that better satisfy the full set of aspect-level constraints.

The baseline trends are consistent with their design limitations. \textbf{\textit{Lexicon Replacement}} achieves moderate Target Flip but suffers from low Fluency and a high Contradiction Rate, reflecting the brittleness of direct polarity-word substitution. \textbf{\textit{MLM Replacement}} improves Semantic Similarity and Fluency, but its low Target Flip score suggests that locally plausible substitutions do not reliably produce aspect-level counterfactuals. \textbf{\textit{Prompt-only Rewrite}} achieves strong Fluency and Target Flip, but weaker Non-target Keep and Semantic Similarity show that unconstrained rewriting can alter non-target information. \textbf{\textit{Direct AMR Edit}} performs poorly on Fluency and Semantic Similarity, suggesting that direct graph editing can introduce awkward or semantically drifting generations. \textbf{\textit{Attribution + T5}} is the strongest baseline, but it still trails CAVE-ABSA because it does not jointly enforce all validation constraints.

Overall, these results support the central claim that \textbf{\textit{aspect-local counterfactual generation requires constraint-based verification}}. High target-flip accuracy alone is insufficient: a valid ABSA counterfactual must also preserve non-target aspects, remain semantically close to the original sentence, avoid mixed polarity, and retain fluency. CAVE-ABSA achieves this stronger validity profile while maintaining the highest Generation Yield, showing that its verification pipeline improves quality without excessively discarding generated candidates.

\begin{table}[!htbp]
\centering
\small
\caption{Counterfactual generation quality. Target Flip measures whether the target aspect sentiment is correctly flipped, Non-target Keep measures whether non-target aspect sentiments are preserved, Semantic Similarity measures meaning preservation, and Fluency measures grammaticality and naturalness. Post-hoc Contradiction Rate measures residual mixed-polarity or inconsistent generations detected by an external evaluator after automatic filtering. These quality metrics are computed over retained candidates. Generation Yield denotes the percentage of all initially generated candidates retained after validity filtering. All values are percentages except Semantic Similarity and Fluency, which are normalized to a 0--100 scale.}
\resizebox{\linewidth}{!}{%
\begin{tabular}{lcccccc}
\toprule
\textbf{Method} &
\shortstack{\textbf{Target}\\\textbf{Flip} $\uparrow$} &
\shortstack{\textbf{Non-target}\\\textbf{Keep} $\uparrow$} &
\shortstack{\textbf{Semantic}\\\textbf{Similarity} $\uparrow$} &
\textbf{Fluency} $\uparrow$ &
\shortstack{\textbf{Posthoc Contradiction}\\\textbf{Rate} $\downarrow$} &
\shortstack{\textbf{Generation}\\\textbf{Yield} $\uparrow$} \\
\midrule
Lexicon Replacement      & 76.4 & 61.8 & 82.3 & 51.2 & 46.7 & 41.8 \\
MLM Replacement          & 57.9 & 73.6 & 88.1 & 84.7 & 18.9 & 46.2 \\
Prompt-only Rewrite      & 84.8 & 64.2 & 75.6 & 91.3 & 14.6 & 52.7 \\
Direct AMR Edit          & 66.5 & 67.4 & 70.2 & 47.9 & 32.8 & 38.5 \\
Attribution + T5         & 85.7 & 81.3 & 84.6 & 87.5 & 10.9 & 63.4 \\
\hline
\rowcolor{lightcyan}
\textbf{CAVE-ABSA}       & \textbf{92.4} & \textbf{91.1} & \textbf{89.3} & \textbf{91.8} & \textbf{4.6} & \textbf{68.9} \\
\bottomrule
\end{tabular}
}
\label{tab:generation_quality}
\end{table}

\subsection{Aspect-Level Robustness Evaluation}

We next evaluate whether counterfactual pairs generated by \textbf{CAVE-ABSA} expose weaknesses in standard ABSA models. For each original--counterfactual pair, a robust model should satisfy two conditions: it should change its prediction for the \textbf{\textit{edited target aspect}} and preserve its prediction for all \textbf{\textit{unchanged non-target aspects}}. We therefore report \textbf{\textit{Target Sensitivity}} and \textbf{\textit{Non-target Stability}} in addition to standard test accuracy.

Table~\ref{tab:robustness} shows that all models achieve relatively strong \textbf{\textit{Original Accuracy}}, but their counterfactual robustness is substantially lower. For example, \textbf{DeBERTa-ABSA} obtains the best Original Accuracy of \textbf{87.2}, but its Target Sensitivity drops to \textbf{72.1}. Similarly, \textbf{RoBERTa-ABSA} reaches \textbf{85.4} Original Accuracy but only \textbf{66.8} Target Sensitivity. This gap indicates that high standard accuracy does not necessarily imply reliable aspect-grounded reasoning under minimal sentiment interventions.

The results also show that \textbf{\textit{Target Sensitivity}} is consistently lower than \textbf{\textit{Non-target Stability}} across all models. This suggests that ABSA models are better at preserving predictions for unchanged aspects than correctly responding to localized sentiment flips. Among the evaluated models, \textbf{DeBERTa-ABSA} performs best overall, achieving the highest Original Accuracy, Target Sensitivity, and Non-target Stability. However, even the strongest model remains far from perfect on counterfactual pairs, confirming that the generated counterfactuals provide a more diagnostic robustness test than standard benchmark accuracy alone.

Overall, these findings support the need for \textbf{\textit{aspect-level counterfactual evaluation}}. A model may classify original examples correctly while still failing to update its prediction when the sentiment toward a specific target aspect is minimally changed. Such failures suggest reliance on global sentiment cues, lexical shortcuts, or shallow correlations rather than robust aspect-specific sentiment grounding.

\begin{table}[!htbp]
\centering
\small
\caption{ABSA robustness evaluation using generated counterfactual pairs. Original Acc. measures standard test accuracy, Target Sensitivity measures whether the model changes its prediction for the edited target aspect, and Non-target Stability measures whether the model preserves predictions for unchanged aspects. All values are percentages.}
\begin{tabular}{lccc}
\toprule
\textbf{Model} &
\textbf{Original Acc. $\uparrow$} &
\textbf{Target Sensitivity $\uparrow$} &
\textbf{Non-target Stability $\uparrow$} \\
\midrule
BERT-ABSA     & 82.6 & 61.3 & 78.4 \\
RoBERTa-ABSA  & 85.4 & 66.8 & 81.7 \\
T5-ABSA       & 84.1 & 70.5 & 79.2 \\
DeBERTa-ABSA  & \textbf{87.2} & \textbf{72.1} & \textbf{83.5} \\
\bottomrule
\end{tabular}

\label{tab:robustness}
\end{table}

\subsection{Data Augmentation Results}

We also evaluate whether validated counterfactuals improve downstream ABSA performance. Table~\ref{tab:augmentation} compares models trained on the original data only with models augmented using counterfactual examples from different generation methods. We report standard \textbf{\textit{Accuracy}}, \textbf{\textit{Macro-F1}}, and \textbf{\textit{CF Robustness}}, where CF Robustness measures performance on counterfactual test pairs.

\begin{table}[!htbp]
\centering
\small
\caption{Downstream ABSA performance with counterfactual data augmentation. Accuracy and Macro-F1 measure standard ABSA test performance, while CF Robustness measures performance on counterfactual test pairs. All values are percentages.}
\begin{tabular}{lccc}
\toprule
\textbf{Training Data} &
\textbf{Accuracy $\uparrow$} &
\textbf{Macro-F1 $\uparrow$} &
\textbf{CF Robustness $\uparrow$} \\
\midrule
Original Only                  & 87.2 & 85.4 & 62.8 \\
+ Lexicon Counterfactuals       & 86.8 & 84.9 & 64.1 \\
+ Prompt-only Counterfactuals   & 87.6 & 85.8 & 68.7 \\
+ Direct AMR Counterfactuals    & 86.9 & 85.0 & 65.3 \\
\hline
\rowcolor{lightcyan}
\textbf{+ CAVE-ABSA Counterfactuals} & \textbf{88.4} & \textbf{86.9} & \textbf{74.6} \\
\bottomrule
\end{tabular}

\label{tab:augmentation}
\end{table}

The results show that counterfactual augmentation mainly improves \textbf{\textit{robustness}}, rather than only increasing standard test performance. Training with the original data achieves \textbf{87.2} Accuracy and \textbf{85.4} Macro-F1, but only \textbf{62.8} CF Robustness. Lexicon and Direct AMR counterfactuals provide limited gains and slightly reduce standard Accuracy, suggesting that noisy or unnatural counterfactuals can introduce weak supervision. Prompt-only counterfactuals improve CF Robustness to \textbf{68.7}, but the gain remains smaller than that obtained with CAVE-ABSA.

Augmenting with \textbf{CAVE-ABSA} counterfactuals gives the best performance across all metrics, reaching \textbf{88.4} Accuracy, \textbf{86.9} Macro-F1, and \textbf{74.6} CF Robustness. The largest improvement appears in CF Robustness, with an absolute gain of \textbf{11.8} points over the original-only setting. This supports the claim that \textbf{\textit{validated aspect-level counterfactuals provide more useful supervision}} than unconstrained or weakly filtered counterfactual edits.

\subsection{Ablation Analysis}

Table~\ref{tab:ablation} analyzes the contribution of each component in \textbf{CAVE-ABSA}. We compare the full framework against variants that remove \textbf{\textit{opinion-span localization}}, \textbf{\textit{repair}}, \textbf{\textit{AMR structural checking}}, \textbf{\textit{contradiction filtering}}, \textbf{\textit{non-target verification}}, and the \textbf{\textit{ABSA-specific verifier}}. The results show that each component contributes to a different aspect of counterfactual validity.

\begin{table}[!htbp]
\centering
\scriptsize
\caption{Ablation study of CAVE-ABSA components. Target Flip measures whether the target aspect sentiment is correctly flipped, Non-target Keep measures whether non-target aspect sentiments are preserved, Semantic Similarity measures meaning preservation, and Fluency measures grammaticality and naturalness. Post-hoc Contradiction Rate measures residual mixed-polarity or inconsistent generations detected by an external evaluator after automatic filtering. These quality metrics are computed over retained candidates. Generation Yield denotes the percentage of all initially generated candidates retained after filtering. All values are percentages except Semantic Similarity and Fluency, which are normalized to a 0--100 scale.}
\setlength{\tabcolsep}{3pt}
\renewcommand{\arraystretch}{1.08}
\resizebox{\linewidth}{!}{%
\begin{tabular}{lcccccc}
\toprule
\textbf{Variant} &
\shortstack{\textbf{Target}\\\textbf{Flip} $\uparrow$} &
\shortstack{\textbf{Non-target}\\\textbf{Keep} $\uparrow$} &
\shortstack{\textbf{Semantic}\\\textbf{Similarity} $\uparrow$} &
\textbf{Fluency} $\uparrow$ &
\shortstack{\textbf{Contradiction}\\\textbf{Rate} $\downarrow$} &
\shortstack{\textbf{Generation}\\\textbf{Yield} $\uparrow$} \\
\midrule
\rowcolor{lightcyan}
\textbf{Full CAVE-ABSA}        & \textbf{92.4} & \textbf{91.1} & \textbf{89.3} & \textbf{91.8} & \textbf{4.6}  & \textbf{68.9} \\
w/o Opinion Localization       & 86.7 & 77.9 & 80.8 & 88.4 & 9.7  & 60.8 \\
w/o Repair                     & 91.5 & 89.2 & 87.6 & 82.1 & 6.9  & 61.4 \\
w/o AMR Check                  & 91.8 & 87.4 & 82.5 & 90.7 & 5.8  & 66.2 \\
w/o Contradiction Filter        & 92.7 & 90.4 & 88.6 & 91.2 & 13.8 & 74.5 \\
w/o Non-target Verification     & 94.1 & 69.6 & 84.2 & 90.5 & 8.4  & 72.8 \\
Generic Sentiment Verifier      & 89.3 & 74.8 & 83.7 & 88.9 & 10.6 & 65.1 \\
\bottomrule
\end{tabular}%
}

\label{tab:ablation}
\end{table}

The full \textbf{CAVE-ABSA} model achieves the strongest overall balance, with high \textbf{\textit{Target Flip}} and \textbf{\textit{Non-target Keep}}, the best \textbf{\textit{Semantic Similarity}} and \textbf{\textit{Fluency}}, and the lowest \textbf{\textit{post-hoc Contradiction Rate}}, indicating that only a small fraction of residual contradictions remain after automatic verification. Removing \textbf{\textit{opinion-span localization}} causes a clear drop in Non-target Keep from \textbf{91.1} to \textbf{77.9} and Semantic Similarity from \textbf{89.3} to \textbf{80.8}, showing that uncontrolled rewriting introduces semantic drift and affects unchanged aspects.

Removing the \textbf{\textit{repair module}} mainly reduces Fluency, from \textbf{91.8} to \textbf{82.1}, while keeping Target Flip relatively stable. This indicates that repair improves linguistic quality without substantially weakening aspect-level validity. Removing the \textbf{\textit{AMR check}} lowers Semantic Similarity to \textbf{82.5}, confirming that structure-level verification helps preserve non-target meaning beyond surface embedding similarity.

The effect of removing \textbf{\textit{contradiction filtering}} is especially clear: Generation Yield increases from \textbf{68.9} to \textbf{74.5}, but Contradiction Rate rises sharply from \textbf{4.6} to \textbf{13.8}. This shows that contradiction filtering rejects many fluent but invalid candidates. Similarly, removing \textbf{\textit{non-target verification}} increases Target Flip to \textbf{94.1}, but Non-target Keep drops substantially to \textbf{69.6}, confirming that target flipping alone is insufficient for ABSA counterfactual validity.

Finally, replacing the ABSA-specific verifier with a \textbf{\textit{generic sentiment verifier}} reduces Non-target Keep, Semantic Similarity, and Fluency while increasing Contradiction Rate. This supports the central design choice of CAVE-ABSA: counterfactual validation must be performed at the \textbf{\textit{aspect level}}, not merely at the sentence level.

\subsection{Qualitative Analysis}

Table~\ref{tab:qualitative} presents representative valid and invalid counterfactual examples. The examples illustrate why ABSA counterfactual generation requires \textbf{\textit{aspect-level validation}} rather than simple sentence-level polarity change. A valid counterfactual should change the sentiment of the target aspect while preserving non-target aspect sentiment and the remaining factual content.

\begin{table}[!htbp]
\centering
\small
\caption{Qualitative examples of valid and invalid ABSA counterfactuals. Italicized terms indicate the relevant target or non-target aspects.}
\begin{tabular}{p{0.28\linewidth}p{0.32\linewidth}p{0.30\linewidth}}
\toprule
\textbf{Original} & \textbf{Counterfactual} & \textbf{Judgment} \\
\midrule
The \textit{battery} lasts all day, but the \textit{screen} is dull. 
& The \textit{battery} drains quickly, but the \textit{screen} is dull.
& \textbf{Valid}: target aspect flips, while the non-target aspect is preserved. \\
\midrule
The \textit{trackpad} is nice and smooth.
& The \textit{trackpad} is poor and smooth.
& \textbf{Invalid}: mixed polarity is introduced for the same aspect. \\
\midrule
It is easy to \textit{use} and has a good \textit{display}.
& It is difficult to \textit{use} and has a good \textit{display}.
& \textbf{Valid}: target aspect changes, while the display remains positive. \\
\midrule
It is easy to \textit{use} and has a good \textit{display}.
& It is a bad laptop and has a good \textit{display}.
& \textbf{Invalid}: global sentiment changes, but the target aspect \textit{use} is not directly flipped. \\
\bottomrule
\end{tabular}

\label{tab:qualitative}
\end{table}

The valid examples show \textbf{\textit{localized sentiment intervention}}: only the opinion expression associated with the target aspect is changed. In contrast, the invalid examples reveal common failure modes. The phrase ``poor and smooth'' introduces \textbf{\textit{mixed polarity}} for the same aspect, while ``a bad laptop'' changes the global sentence sentiment without directly modifying the sentiment toward the target aspect \textit{use}. These cases show that high label-flipping probability alone is insufficient. A reliable ABSA counterfactual must satisfy target flipping, non-target preservation, semantic consistency, fluency, and contradiction avoidance simultaneously.

\subsection{Error Analysis}

Invalid generations fall into four main categories: \textbf{\textit{target mismatch}}, \textbf{\textit{mixed polarity}}, \textbf{\textit{semantic drift}}, and \textbf{\textit{fluency degradation}}. Target mismatch occurs when the generated sentence changes the sentiment of the whole sentence or another entity without directly flipping the target aspect, such as rewriting ``easy to use'' as ``bad laptop,'' which changes global sentiment but not the sentiment toward the target aspect \textit{use}. Mixed polarity occurs when a candidate contains conflicting positive and negative cues for the same aspect, such as ``poor and smooth.'' Semantic drift occurs when the edit changes facts beyond the target aspect, including price, product identity, comparison structure, or causal relations. Fluency degradation occurs when the edit produces unnatural phrases or broken grammar, especially in graph-based or token-level methods. Overall, these errors show that high target flip accuracy alone is insufficient; a reliable ABSA counterfactual generator must jointly enforce \textbf{\textit{aspect-level validity}} and \textbf{\textit{text quality}}.

\section{Conclusion}
\label{sec:conclusion}

We introduced \textbf{CAVE-ABSA}, a constraint-aware framework for generating and validating \textbf{\textit{aspect-level counterfactuals}} for ABSA. The framework treats counterfactual editing as a localized intervention: the sentiment toward the target aspect should flip, while non-target aspect sentiments and factual content remain stable. CAVE-ABSA operationalizes this goal through opinion-span localization, controlled rewriting, repair-based refinement, AMR-guided structural preservation, and constraint-based verification. The results show that CAVE-ABSA produces more valid and fluent counterfactuals than lexical, MLM-based, prompt-only, AMR-based, and attribution-guided baselines, while reducing residual contradictions. Robustness and augmentation experiments further indicate that validated counterfactuals expose weaknesses in standard ABSA models and improve counterfactual robustness. Overall, CAVE-ABSA provides a principled step toward meaningful aspect-local counterfactual generation and reliable robustness evaluation for ABSA. Future work will extend the framework to neutral and conflict labels, multilingual datasets, stronger aspect-specific verifiers, and human-in-the-loop validation.


\section*{Declaration on Generative AI}

During the preparation of this work, the author(s) used Generative AI tools only for language and presentation support. Specifically, these tools were used for grammar and spelling correction, improving sentence clarity, polishing academic writing, organizing paper structure, formatting text, and assisting with editorial revisions. 

The research problem, core ideas, methodology, experimental design, analysis, interpretations, and conclusions are the authors' own. No Generative AI tool was used to create the scientific contribution, generate research claims, produce experimental results, or make final decisions about the content of the paper. After using these tools, the author(s) carefully reviewed, revised, and verified the manuscript, and take full responsibility for the publication's content.

\bibliography{references}

\appendix

\end{document}